\documentclass[lettersize,journal]{IEEEtran}
\usepackage{amsmath,amsfonts}
\usepackage{algorithmic}
\usepackage{algorithm}
\usepackage{array}
\usepackage[caption=false,font=normalsize,labelfont=sf,textfont=sf]{subfig}
\usepackage{textcomp}
\usepackage{stfloats}
\usepackage{url}
\usepackage{verbatim}
\usepackage{graphicx}
\usepackage{cite}
\usepackage{amssymb} 
\begin{document}

\title{Deep Reinforcement Learning Based Multi-Access Edge Computing Schedule for Internet of Vehicles}

\author{Xiaoyu~Dai,~\IEEEmembership{Student Member,~IEEE,}
        Kaoru~Ota,~\IEEEmembership{Member,~IEEE,}
        and~Mianxiong~Dong,~\IEEEmembership{Member,~IEEE}
\thanks{
Xiaoyu Dai, Kaoru Ota and Mianxiong Dong are with the Department of Sciences and Informatics, Muroran Institute of Technology, Japan  (e-mail: 20043036@mmm.muroran-it.ac.jp; ota@csse.muroran-it.ac.jp; mx.dong@csse.muroran-it.ac.jp).}
}

\markboth{Journal of \LaTeX\ Class Files,~Vol.~14, No.~8, August~2021}%
{Shell \MakeLowercase{\textit{et al.}}: A Sample Article Using IEEEtran.cls for IEEE Journals}


\maketitle

\begin{abstract}
As intelligent transportation systems been implemented broadly and unmanned arial vehicles (UAVs) can assist terrestrial base stations acting as multi-access edge computing (MEC) to provide a better wireless network communication for Internet of Vehicles (IoVs), we propose  a UAVs-assisted approach to help provide a better wireless network service retaining the maximum Quality of Experience (QoE) of IoVs on the lane. In the paper, we present a Multi-Agent Graph Convolutional Deep Reinforcement Learning (M-AGCDRL) algorithm which combines local observations of each agent with a low-resolution global map as input to learn a policy for each agent. The agents can share their information with others in graph attention networks, resulting in an effective joint policy. Simulation results show that the M-AGCDRL method enables a better QoE of IoTs and achieves good performance.
\end{abstract}


\section{Introduction}
\IEEEPARstart{R}{ecently}, Internet of Vehicles has been playing an essential role in creating a safer and more intelligent transportation system in the upcoming 5G (the fifth generation) network. 
The concept of IoVs origins from IoTs (Internet of Things), which provides network connection service between the vehicles on the road with the other vehicles, pedestrians, and roadside infrastructures. The massive implementation of IoVs greatly improves the Quality of Experience of the drivers and the transportation efficiency. 

IoVs support five types of network communication, they are Vehicle to Infrastructure (V2I), Vehicle to Vehicle (V2V), Vehicle to Pedestrian (V2P), etc.(or broadly the V2X communication) \cite{Lyu2018}, \cite{Xu2018}. With the autonomous vehicles been carried out in more and more countries, the V2X will be of more significance in the future. However, with the dramatic growth of the data traffic and bandwidth limitation of the Internet, there are some drawbacks that can not be neglected, since the enormous traffic data corresponds to enormous network transmission cost. One of the drawbacks is that current network bandwidth is not enough to support too many network connection devices, especially when the traffic is heavy. Moreover, the V2X requires a very low latency and a relatively high transmission to ensure its regular running, otherwise it will be rather dangerous.

In order to ensure a extensive area coverage, a low latency, and a relatively high transmission speed for IoVs, some methods have been put forward. Companies like Apple and Google have developed applications, Apple CarPlay and Google Android Auto to help drivers plan route referring to real-time traffic condition by connecting iPhone or Android Phone to the vehicle. 

Some applications support offline route planning considering network conditions during a traffic jam, or somewhere remote away from the BS. The defect is obvious since offline route planning lacks plenty of real-time details, which will largely degrade the users' experience. \cite{Chen2016}, \cite{Zhang2017} come up with the long term evolution (LTE)-V wireless standard to improve capacity to improve network performance. In LTE-V, every vehicle shares the same network resource. The two modes of LTE-V, LTE-V-direct and LTE-V-cell cooperate to provide wireless network service to vehicles on the roads.

Nevertheless, the LTE base station is stationary, which brings significant challenges to satisfy the high data rate and low latency, especially in dense IoVs scenarios. Besides, it costs a lot to construct a BS specifically in remote areas. Apart from that, it is not necessary or sometimes not possible to set up a new terrestrial base station timely, especially during the aftermath of natural disasters. 

As wireless communication and smart car technologies become more sophisticated, IoV has grabbed more and more attention over the past few years. Not only for manual driving, but also for autonomous vehicles, the network is widely recognized as possessing considerable value in improving vehicular safety, mobility, sustainability and efficiency. Meanwhile, vehicular networks also face great challenges mainly because of the emergence of those latency-sensitive and computation-intensive vehicular applications, due to the limited spectrum resources and limited computation resources.

Recently, flying MEC devices such as UAVs have been served as promising assistants helping provide better wireless networks, especially when equipped with communication transceivers. Owing to their low cost and high flexibility, the UAVs are able to provide wireless connectivity for IoVs assisting LTE network provided by the Base Station. 

Deploying some UAVs can perfectly accomplish the task as a temporary wireless base station. Compared to these conventional methods, UAV-enabled wireless communications have emerged as a promising way to help ensure stable network service\cite{Liu2018}, \cite{Zeng2016}. Acting as flying MEC devices, those UAVs can significantly improve network capacity with their high mobility and flexible deployment. Apart from that, their cost is relatively low; therefore, it can be simply implemented. By associating with UAVs, better network service can be provided for devices in vehicular network, which leads to an upgrade of QoE for drivers in IoVs.

In spite of the leverage of MEC servers and devices, it is stil very challenging to dynamically support those vehicle application for network service provider, who should deploy the MEC server in advance, otherwise it would be difficult to meet the dynamic resource requirements of the vehicular applications.
Technical challenges include the location and number of MEC servers, and the way multidimensional resources allocate across vehicles.
From driver's perspective, the key is to decide if, when, and how many tasks should be offloaded to the MEC server to meet the QoE requirements.

Although MEC-enabled wireless communication shows its superiority on the network performance of IoVs, it still faces a lot of challenges. For instance, because of the high speed of vehicles, it is hard to decide when and where to deploy the UAVs so that we can not only satisfy drivers' needs but also minimize the waste of UAV resources. Besides, considering about obtaining the best QoE continuously, how to schedule the UAVs trajectories is of extraordinary significance. However, to schedule multiple UAVs has been a difficult problem, according to recent research works. Most researchers choose only one UAV to solve their problems, whereas in practical application scenarios, one UAV is far from enough. 

To tackle these challenges, deep DRL was addressed in this paper. DRL combines DL(deep learning) and RL(reinforcement learning) to create algorithms, which can be applied in many fields such as robotics, video games, and transportation and so on. Frameworks based on DRL is capable of many unsolvable problems due to its unsupervised learning. With the implementation of DRL, we can find the solution to achieve the best QoE by enormous repeated training. Meanwhile, for a problem without prior probability, deep reinforcement learning has the ability to solve it.

The contributions of this paper can be summarized as the below.

1) We proposed a Multi-Agent Graph Convolutional Deep Reinforcement Learning (M-AGCDRL) algorithm, which applies multiple UAVs working as MEC to assist the BS, aiming to achieve the utmost capacity of network. Those UAVs can substantially relieve the burden caused by the excess amount of devices in vehicular network on the lane.

2) The proposed algorithm combined the graph convolutional networks and actor-critic architecture to maximum the QoE for the drivers. 

3) Simulation results show that the proposed algorithm can improve the QoE for drivers.

The rest of this paper is organized as follows. Related works are discussed in Section II. In Section III, the system model is described. In Section IV, we describe the details of the proposed algorithm. After that, the simulation results are shown in Section V. At last, our conclusions are drawn in Section VI.

\section{Related Works}
In MEC-assisted vehicular networks, the vehicular applications can get access to servers through different wireless communications mainly including dedicated short-range communication (DSRC), Wi-Fi, and cellular. 
While the requirement for computing/storage resources is strong, vehicles perform offloading with MEC servers with a shorter response latency through reducing data transfer between MEC servers and cloud servers.



In \cite{Liu2018}, an energy-efficient control algorithm was proposed for UAVs coverage and connectivity based on deep reinforcement learning to control the UAVs with minimizing their energy consumption. Coverage, fairness, and energy consumption are optimized as simulation shows. 

An overview of UAV-assisted wireless network through the basic networking structure and main channel characteristics was proposed in \cite{Zeng2016}. The paper discussed communications not only between UAVs and the ground, but also between UAVs themselves. Besides, due to short-range line-of-sight (LoS) links, UAV-assisted wireless communications tend to have better communication channels. Three aspects of UAV trajectory design, UAV energy consumption and multiple-input multiple-output (MIMO) communications among UAVs are considered. 

Joze Kosmerl and Andrej Vilhar \cite{Kosmerl2014} proposed a method for calculating the optimal proprieties and coordinates of aerial stations and terrestrial base stations, which shows advances in low altitude platforms (LAPs) in high bandwidth utilization, especially when terrestrial base stations are under destroyed after a natural disaster, especially for catastrophes. The algorithm can deploy minimum number of UAVs to fully cover the target area.

AL-Hourani et al. \cite{AlHourani2014} present an analytical solution for LAP altitude optimization. An optimal altitude is calculated for providing maximum radio coverage on the ground. Apart from this, a closed-form formula is proposed to predict the probability of geometrical LoS from the platform to a ground user. Considering the fact that air- to-ground radio channel behaves randomly, it still requires further investigation.

Alzenad et al. \cite{Alzenad2017} proposed an optimal deployment algorithm for UAV-assisted Base Station wireless network, and the algorithm provides as many the number of users the network can cover at the least consumption of transmit power. The three-dimensional deployment algorithm they have proposed decouples the problem into UAV's optimal vertical position and horizontal coordinate separately without any optimal loss.

In \cite{Mozaffari2016}, the authors proposed a framework to obtain the optimal three-dimensional positions of multiple UAVs, where they can achieve a maximum ground coverage and minimum transmit power. In this paper, circle packing theory is made use of to maximize the zone that UAVs can cover. In the meantime, the least UAVs are required to assure goal coverage for the determined ground district.



According to \cite{Zeng2017}, Yong Zeng and Rui Zhang supposed that the UAV would be flying at a consistent altitude. Under this premise, they developed a model through optimizing the flying distance and speed of UAV to make the best of communication energy efficiency under a practical circular UAV trajectory. 

In \cite{Hu2019}, a sense-and-send protocol was come up with, the protocol aims at coordinating multiple UAVs to carry out a real-time sense and send task. Then an enhanced DRL algorithm for multiple UAVs is proposed to solve the decentralized multi-agent trajectory design problem. This paper proved that DRL could be applied to UAV-assisted wireless communications.

A novel framework proposed in \cite{Mozaffari2017} makes full use of UAV acting as an aerial station to maximize the throughput of users. Such UAV-based wireless network systems are optimized with regard to the average bits between UAV and users as well as the duration UAVs hover. UAVs' 3-dimensional deployment was attained with the help of the circle packing theory.


To make an interference-aware UAV trajectory scheme, authors in \cite{Challita2019} proposed a DRL algorithm to both maximize energy efficiency and minimize wireless latency. Also, to minimize the computational complexity of the proposed algorithm, an upper and a lower bound for the flying altitude are set in advance.

An UAV-aided cellular communication framework is proposed in \cite{Lu2018} to assist cellular communications hold out against smart jamming. With the help of deep reinforcement learning, the scheme can achieve a much less error rate through training the previous anti-jamming relay data.

Machine learning techniques are also utilized to solve UAV trajectory design and obstacle avoidance problems, authors in \cite{Yijing2017} applied neural network to calculate Q-value so that they can avoid a low-quality trajectory and massive computation of matrix while ensuring the precision. 

Mozaffari et al. \cite{Mozaffari2019} developed a new concept of 3D cellular networks integrating base stations and cellular-connected users through UAV. Utilizing this framework, a method for the minimum number of UAVs converging most is proposed. Simulation shows the proposed latency-optimal cell association can improve the spectral efficiency of a UAV-assisted communication.


In \cite{Lyu2019}, Feng Lyu et al. developed an Online UAV Scheduling (OUS) scheme aiming at obtaining satisfied throughout performance. UAVs are scheduled and managed overall to assure smooth connections between UAVs and users. Results show that the scheme can efficiently respond to the QoS demand through sending as few as UAVs intelligently.

Yuanwei Liu with his partners \cite{Liu2019} proposed a novel framework for UAV-assisted communications with a large number of access capabilities. The framework is supported by no-orthogonal multiple access. They applied stochastic geometry to create the coordinates of UAVs and users on the ground, then analyzed the model with single UAV and multiple UAVs separately.

\begin{figure*}
    \centering
    \includegraphics[scale=0.58]{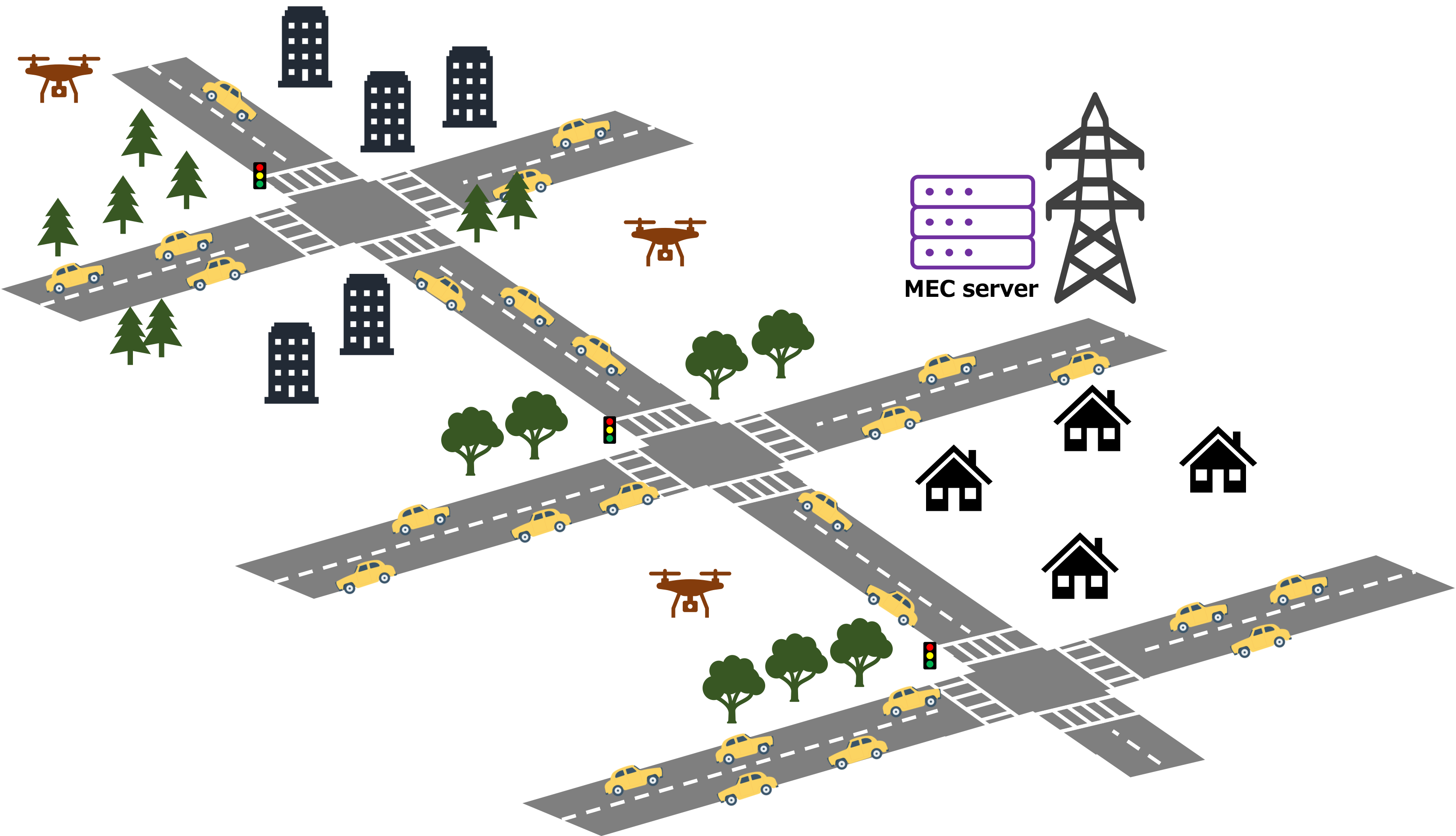}
    \caption{The MEC-enabled wireless communications architecture.}
    \label{Fig. 1: UAV-enabled wireless communications architecture}
\end{figure*}

\section{System Model}
The detailed system model in 3-dimensional scenario is described in this section. We create an environment in which the BS can fully cover IoVs in the area, and multiple UAVs can be sent alternatively to assist the BS. The UAV-assisted BS wireless network aims at providing a better network service for drivers on the road. In this paper, we consider a fixed down-link transmission rate ${T_r}$ is in demand for each IoV. In order to maximize the QoE of IoVs, multiple UAVs are utilized to support the BS when the network resource provided by the BS is not abundant enough.

\subsection{System Description}
As is illustrated in Fig. 1, we consider a UAV-enabled wireless communications with a BS and $N$ UAVs. We consider there is one BS in the environment. Assume that the BS acts as the origin of the system coordinate, and the height of the BS is $H_{BS}$. The set of the UAVs and vehicles are denoted by $\mathcal{N} = \{1, 2, ..., N\}$ and $\mathcal{M} = \{1, 2, ..., M\}$, respectively. Supposing the coordinate of each UAV as ${U_n} = [{x_{{u_n}}}(t),{y_{{u_n}}}(t)]$, where ${x_{{u_n}}}(t)$ and ${y_{{u_n}}}(t)$ are the coordinate of UAV $N$ at time t. The height of each UAV is set stationary as $H_{u}$ in this paper. Each vehicle coordinates at ${V_n} = [{x_{{v_m}}}(t),{y_{{v_m}}}(t)]$, where ${x_{{v_m}}}(t)$ and ${y_{{v_m}}}(t)$ are the coordinate of vehicle $M$ at time t. As the BS offers the LTE service whose transmission rate has a lot to do with the distance between the BS and vehicles. The distance from the BS to each vehicle is expressed by $M$ with $D$, given by
\begin{equation}
    D = \sqrt {{H_u}^2 + {x_{v_m}}^2 + {y_{{v_m}}}^2}\text{.}
\end{equation}
which reflects the straight line distance between them.

Denote all the vehicles associate to the UAV are distributed with equal bandwidth. Note the total bandwidth of the BS as $W_{BS}$. The BS can be separated into at most $X$ sub channels. Therefore, the download throughput of the vehicle M from the BS is
\begin{equation}
    {B_{{v_m}}} = \left\{ \begin{array}{*{20}{c}}
\frac{W_{BS}}{M}\log_2(1 + \frac{p_{t_m}p_{c_m}}{\frac{W_{BS}M{n_0}})} & M < X\\
\frac{W_{BS}}{M}\log_2(1 + \frac{{{p_{{t_m}}}{p_{{c_m}}}}}{{\frac{{{W_{BS}}}}{X}{n_0}}}) & M \ge X & 
\end{array} \right.\text{.}
\end{equation}
where $p_{{t_i}}$ denotes the transmission power, $p_{{c_i}}$ denotes the channel power and $n_0$ means the background density of the additive white Gaussian noise (AWGN).
Besides, UAV $M$'s transmission rare can be signified as
\begin{equation}
    {R_{{v_m}}} = {B_{{v_m}}}{\log _2}\left[ {1 + \frac{{{p_{{t_m}}}{p_{{c_m}}}}}{{{\sigma ^2}}}} \right]\text{.}
\end{equation}
where $\sigma ^2 = {B_{{v_m}}}{n_0}$.

Since the UAVs are flying in a 3-D space, the probability of LoS connections for the UAVs is higher than on terrestrial conditions. The LoS probability can be denoted as
\begin{equation}
    P_{LoS}(\theta _{k_{n}})=b_{1}(\frac{180}{\pi }\theta _{k_{n}}-\zeta )^{b_{2}}
\end{equation}
where $\theta _{k_{n}}(t)=sin^{-1}\left [ \frac{h_{n}(t)}{d_{k_{n}}(t)} \right ]$ is the elevation angle between UAV $n$ and the vehicle $M$. $b_{1}$, $b_{2}$ and $\sigma $ are environment-determined constants. Besides, the Non-Line-of-Sight (NLoS) probability is given by $P_{NLoS}=1-P_{LoS}$.
From $P_{LoS}$ and $P_{NLoS}$ the channel power from UAV $N$ to UAV $M$ can be denoted as
\begin{equation}
    g_{k_{n}}(t)=K_{0^{-1}}d_{k_{n}}^{-\alpha }(t)\left [ P_{LoS}\mu _{LoS}+P_{NLoS}\mu _{NLoS} \right ]
\end{equation}
where $K_{0}=\left ( \frac{2\pi f_{c}}{c} \right )^{2}$, $\alpha $ is constant which denotes path loss exponen. $\mu _{LoS}$ and $\mu _{NLoS}$ are different attenuation factors considered for LoS and NLoS links, $f_{c}$ is the carrier frequency, $c$ is the speed of light.
As the received interference from UAV to vehicle can be mitigated by assigning different spectrum to each cluster, the received SNR $\Gamma _{v_{m}}(t)$ of vehicle $M$ associated with UAV $N$ at time $t$ can be denoted as
\begin{equation}
    \Gamma _{v_{m}}(t)=\frac{p_{t_{m}}p_{c_{m}}}{\sigma ^{2}}
\end{equation}
Since the transmit rate of the vehicles requirements varies from each other, we denote a SNR threshold $\gamma _{v_{m}}$. If the SNR a vehicle receives doesn't reach this threshold, we consider it as a failed transmission, that is, $\gamma _{v_m} \geqslant \gamma _{v_m}$.

\begin{table*}[h]
\centering
\caption{Notation List}
\label{tab:my-table}
\begin{tabular}{|c|c|c|c|}
\hline
\textbf{Notation}                 & \textbf{Description}  & \textbf{Notation}        & \textbf{Description} \\ \hline
$M$                               & Number of vehicles    & $H_{u}$     & Height of UAVs       \\ \hline
$N$                               & Number of UAVs        & $W_{BS}$     & Total bandwidth of the BS                \\ \hline
$\mathcal{N}$                     & Set of vehicles       & ${B_{{v_m}}}$     & Download throughput from BS               \\ \hline
$\mathcal{M}$                     & Set of UAVs           & $p_{{t_i}}$       & Transmission power              \\ \hline
$D$                               & Distance between BS and vehicle  & $p_{{c_i}}$  & Channel power \\ \hline
$[{x_{{v_m}}}(t),{y_{{v_m}}}(t)]$ & Coordinate of UAV $M$        & $n_0$       & Background density of AWGN     \\ \hline
$[{x_{{u_n}}}(t),{y_{{u_n}}}(t)]$ & Coordinate of UAV $N$        & ${R_{{v_m}}}$    & UAV $M$'s transmission rare         \\ \hline
\end{tabular}
\end{table*}

\subsection{Network Resource Shortage Monitoring}
As the vehicles move fast in and out all the time, the density of the vehicles on each lane changes continuously. We set every each intersection $I_{i}$ and all the lanes $L_{j}$ connected to it as one block. At each time slot $k$, the density of the block is calculated as $D_{k}=\frac{\sum V}{\sum length(L)}$, where $length(R)$ represents the length of the lane $L$ at time $t_{k}$. Given a threshold $\lambda$, when the density of one block exceeds the threshold, network resource shortage is detected, and UAVs are ready to be deployed to reduce the burden on BS.

\subsection{Quality-of-Experience Model}
QoE measures a customer's experience with spectacular service, whether he or she is satisfied or dissatisfied. As customers' experience varies from each other, it is essential to set a standard to evaluate the service provided. Mean opinion score (MOS) is utilized for QoE evaluation, The MOS is represented on a five alternatives where $5 = excellent$, $4 = good$, $3 = fair$, $2 = poor$, $1 = bad$. The definition of QoE is as follows
\begin{equation}
    MO{S_{{v_m}}}(t) = {\lambda _1}MOS_{{v_m}}^{delay}(t) + {\lambda _2}MOS_{{v_m}}^{rate}(t)\text{.}
\end{equation}
where${\lambda _1} + {\lambda _2} = 1$, ${\lambda _1}$ and ${\lambda _2}$ are coefficients.
As the transmission rate is the most important factor, we ignore the delaying time and simplified the MOS model as
\begin{equation}
     MO{S_{{v_m}}}(t) = MOS_{{v_m}}^{rate}(t)\text{.}
\end{equation}
To maximize the sum MOS of each vehicle during the period time $T_p$, which is defined as follows
\begin{equation}
    MO{S_v} = \sum\limits_{t = 0}^{{T_p}} {MO{S_{{v_m}}}(t)}\text{.}
\end{equation}

\subsection{Problem Formulation}
We are aiming at maximizing the MOS of the drivers on the vehicles in a MEC-assisted network, the problem can be formulated as
\begin{equation}
    max MOS_{total}=\sum_{n=1}^{N}\sum_{k_{n}=1}^{K_{n}} \sum_{t=0}^{T_{n}}MOS_{k_{n}}(t)
\end{equation}
which means summarize all the UAVs' MOS during the whole time, and Q-learning is applied to maximize this value in Section IV and Section V.

\begin{figure}[h]
    \centering
    \includegraphics[scale=0.8]{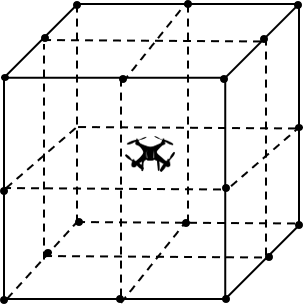}
    \caption{The UAV trajectory design.}
    \label{fig:my_label}
\end{figure}

\subsection{UAV deployment problem}
Applications of Q-learning are always set in a grid world where agents are trained to find the best trajectories to reach the goal. In this paper, each UAV is considered as an agent, and to get the maximum sum of MOS is the final goal. Since Q-learning does not perform well in high dimensional environment because of the restriction of q-table, we attempt to simplify the three dimensional UAV deployment problem by considering the three dimensional world into a finite set of discrete spatial points $S_{p}$ in Fig. 2. At the initial of each time slot, the UAVs can select their flying directions to another 26 points or stay still.

As there are limitations for single-agent DRL framework to solve massive IoVs problems, we adopt a MEC-assisted multi-agent DRL framework in this paper. In this way, we can deploy and plan trajectories for multiple UAVs to help assist the BS with network service.


\subsection{UAV trajectory problem}
In this section, we described the trajectories design of UAV to maintain the QoE of users. As the vehicles move continuously, the UAV' optimal position changes at every cycle. As a result, the UAV needs to travel to offer a better down-link service as the vehicles move. Otherwise, the users’ QoE requirements will not be satisfied.

Conventionally, wireless network tends to deploy more flying BS, the UAVs to serve a more portable network experience. However, considering the high flexibility of UAVs, we can find a trajectory for the UAV to get the optimal MOS without sending more other UAVs. To obtain the trajectories of the UAVs, we assume that the UAV can select its flying directions at the beginning of each cycle, and the vehicles will not roam into the others.

A straightforward method is to decentralize continuous state data. However, without knowing the decentralization granularity in advance, the state may cause a large error in reflecting the actual environment. If we want to ensure accuracy, it may bring about a reduction in the calculation of the matrix.

As shown in Fig. 3, at the beginning of every time slot, each agent carries out an action ${{\rm{a}}_t} \in A$. That is, each agent chooses a direction according to current state, ${{\rm{s}}_t} \in S$ by the policy $J$. Since 27 directions are too much to be calculated by computer, we choose nine points from 27 points as the action space to speed up the convergence speed of  the training. The action space includes nine actions that can be carried out, including the eight corners of the cube in Fig. 2 and stay still.

\subsection{Action Space}
The position of agents are denoted by ${U_n} = [{x_{{u_n}}}(t),{y_{{u_n}}}(t)]$, indicating the horizontal position of each UAV at time slot $t$. The altitude of each UAV is set as $H$, which is decided through UAV deployment algorithm in Section IV. Besides, the altitude of each UAV keep still all the time through the trajectory design. Therefore, the sate space can be denoted as ${{\rm{x}}_{{u_t}}}{\rm{:\{ 0,1,}}...{\rm{,}}{{\rm{X}}_m}{\rm{\} }}$, ${{\rm{y}}_{{u_t}}}{\rm{:\{ 0,1,}}...{\rm{,}}{{\rm{Y}}_m}{\rm{\} }}$, where ${{\rm{X}}_m}$ and ${{\rm{Y}}_m}$ are still the maximum coordinate of the area.

As for a $10\times 10$ two dimensional panel, the state space includes $80$ points in total for UAV to choose. One UAV can only choose to move one step, $\frac{1}{2}$ of the cube's side length in Fig. 2, or stay still. UAVs are initially deployed where the deployment algorithm has decided.


\begin{figure}[h]
    \centering
    \includegraphics[scale=0.4]{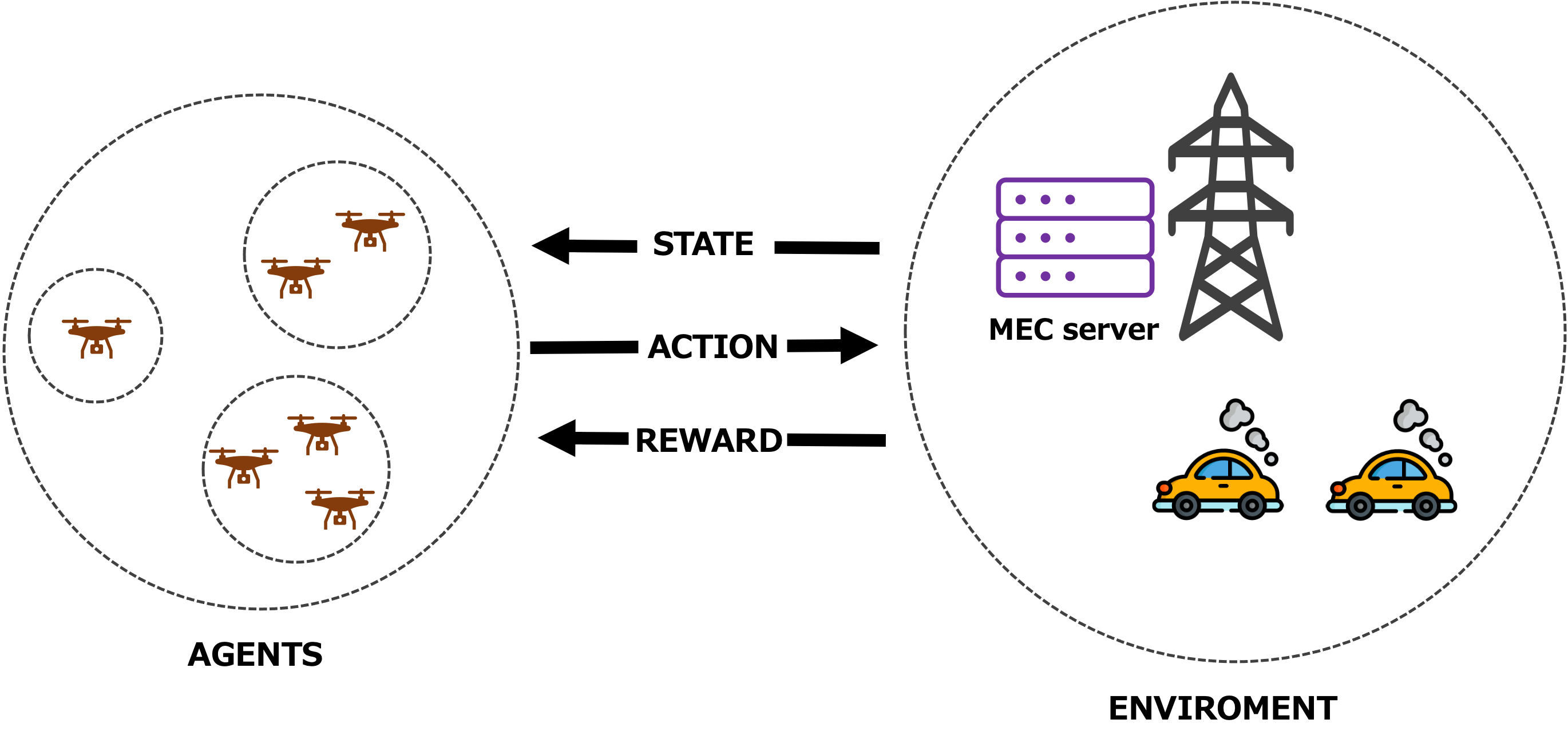}
    \caption{The illustration of the RL system.}
    \label{Fig. 1: UAV-enabled wireless communications architecture}
\end{figure}

\subsection{State Space}
The position of agents are denoted by ${U_n} = [{x_{{u_n}}}(t),{y_{{u_n}}}(t),{h_{{u_n}}}(t)]$, which indicates the position of each UAV in three dimensional space at time slot $t$. The initial altitude of each UAV is set as $H$, which is parallel to the BS horizontally. The sate space can be denoted as ${{\rm{x}}_{{u_t}}}{\rm{:\{ 0,1,}}...{\rm{,}}{{\rm{X}}_m}{\rm{\} }}$, ${{\rm{y}}_{{u_t}}}{\rm{:\{ 0,1,}}...{\rm{,}}{{\rm{Y}}_m}{\rm{\} }}$, where ${{\rm{X}}_m}$ and ${{\rm{Y}}_m}$ are the maximum coordinate of the area.
[x] shows that the further a UAV hovers from the BS, in other words, the closer a UAV hovers to the marginal of the environment, the larger overall throughout can be gained. In this paper, we decide the UAVs' initial position set on the marginal position for each trial.

As for a three dimensional space of $10\times 10\times 4$, the state space includes 158 points in total for UAV to choose. One UAV can only select to move one step, $\frac{\sqrt{2}}{2}$ of the cube's side length in Fig. 2, or stay still. UAVs are initially deployed at the farthest points from the BS, then they are applied with Q-learning algorithm to explore to best initial position to get the maximum MOS.


\subsection{State Transition Mode}
During each time slot $t$, each agent $M$'s state ${s_t}$ transits to state ${s_{t + 1}}$ after action ${a_t}$ is carried out. Action ${a_t}$ is randomly chosen based on the conditional transition probability $p({s_{t + 1}},{r_t}|{s_t},{a_t})$. The Q-learning model aims at achieving maximum long-term rewards ${G_t}$, which can be denoted as follow
\begin{equation}
    {G_t} = E|\sum\limits_{n = 0}^\infty  {{\beta ^n}{r_{t + n}}} |\text{,}
\end{equation}
where $\beta$ is a discount factor.
The problem can be described by a tuple $ < {s_1},...,{s_M},{a_1},...,{a_M},{\rm T},{p_{{r_1}}},...,{p_{{r_M}}},\beta  > $, where
\begin{itemize}
    \item UAVs are the agents.
    \item ${s_1},...,{s_M}$ are state spaces of UAV $M$, indicates the position of UAV $M$.
    \item ${a_1},...,{a_M}$ are the actions consists of nine possible actions $(0,0,0,1)$ means upward back right, $(1,1,1,1)$ means upward back left; $(1,0,0,1)$ represents downward back right, $(0,1,1,1)$ represents downward back left; $(0,1,0,1)$ indicates upward front right, $(1,0,1,1)$ indicates upward front left; $(0,0,1,1)$ denotes downward front right, $(0,0,1,1)$ denotes downward front right; finally, $(0,0,0,0)$ implies stay still. 
    \item ${p_{{r_1}}},...,$ denote the possibility of actions related to rewards for UAV $M$.
    \item $\beta  \in [0,1)$ is the discount factor. It decides how much proportion future rewards account for.
\end{itemize}
We define the reward function as follow,
\begin{equation}
    {R_t} = \left\{ {\begin{array}{*{20}{c}}
{0.8,}&{if{\rm{ }}MO{S_n} < MO{S_{n + 1}}}\\
{ - 0.1,}&{if{\rm{ }}MO{S_n} = MO{S_{n + 1}}}\\
{ - 0.8,}&{if{\rm{ }}MO{S_n} > MO{S_{n + 1}}}
\end{array}} \right.\text{,}
\end{equation}
where ${MO{S_n}}$ and ${MO{S_{n + 1}}}$ represents $t$-th cycle and ${t+1}$-th cycle, respectively. [xx] shows that changing the value of reward does not make difference on algorithm's final result, but influences the convergence rate.
In this paper, we assume that every transmission is successful. The Q-table is iteratively updated during learning in the framework. The BS can inform the UAVs of the rewards of the last cycle, and it is  
\begin{equation}
{Q_{t + 1}}({s_t},{a_t}) \leftarrow (1 - \alpha ){Q_t}({s_t},{a_t}) + \alpha [{R_t} + \beta  \cdot {\max _a}{Q_t}({s_{t + 1}},a)]
\end{equation}

\subsection{UAVs deployment algorithm}
\subsubsection{Single UAV deployment algorithm}
As is mentioned above, Q-learning with single agent in two-dimensional world, the most fundamental DRL algorithm, has been put into various scenarios. We firstly expand the two-dimensional world into three-dimensional world with only one agent, that is, one UAV to assist with the network service.

In the single agent algorithm, there is only one UAV hovering in a constant environment. It observes the sates $s$ and choose an action $a$, which leads to a maximum reward $r$. We take every UAV as an independent agent, and the other UAVs are regarded as part of the environment. Then we can update the q-table according to the following rule,
\begin{equation}
    {Q_{N}}({s_t},{a_t}) \leftarrow {Q_t}({s_t},{a_t}) + \alpha [{R_t} + \beta  \cdot {\max _a}{Q_N}({s_t},a_t)]
\end{equation}
where $\alpha $ is the learning rate.

\subsubsection{Multiple UAVs deployment algorithm}
In the multiple agents algorithms, there are more than one UAVs taking actions simultaneously. In spite of the fact that single agent algorithm takes advantages in smaller state space, it ignores the other agents' action strategies. To solve this problem, we choose an opponent modeling Q-learning to make a better solution.


 The UAVs are deployed randomly close to marginal position initially. Then in every cycle each UAV $N$ choose an action $a_t$ based on the environment and other UAVs' observed frequency distribution in state $s_t$. After every time slot, the q-table will be updated.


The process can be described as follows. Firstly, the positions of UAVs are set to the initial states in environment through the results of deployment. Secondly, the environment observes current states and judges whether the QoE reaches the threshold, if it does, take the history into a success trajectory set; if not, try to reach the threshold until time expires. Then each agent selects an action and takes it, then the state is set to the new state, and the process will be repeated until it reaches the terminal.

\section{Proposed M-AGCDRL algorithm}
In the MEC-assisted wireless network of IoTs, the environment state is subject to dynamic changes at all times due to the infinite channel state. In other words, it is difficult for the agents to allocate resource through successive time slots according to the currently observed environment state. Therefore, in this work, we decide to propose a environment-free DRL algorithm, in which the environment is uncertain. As the size of the state space and the action space is infinite, the algorithms such as strategy gradient, actor criticism, deterministic strategy gradient (DPG) and Deep DPG (DDPG), are often widely used by researchers. In this article, we combine graph convolutional network and actor criticism architecture to solve modeled MDP.

\subsection{The M-AGCDRL architecture}
Our architecture uses deep reinforcement learning structures and graph attention networks to simulate communication. We refer to this architecture as M-AGCDRL. 

Fig. 4 is the schematic diagram of M-AGCDRL. Each agent shared a same structure with the same weight. This makes it possible to promote the setting by duplicating the structure for more agents. As is shown in Fig. 4, the structure is made up of by three parts, which are: a convolutional neural network for extracting features from the input of each agent; a graph attention network for sharing information among agents; an actor criticism algorithm, using to find the best strategy for each agent.

\subsection{CNN and GAT}
Firstly, we extract the feature vector $hc_i$ from $w_i$ with a CNN. The observed value $z_n$ of all agents $N$ at time $t_p$ is input to the network. Then, the feature vector is passed to a GAT and also sent to its neighbour agents.

Then, the graph attention network is applied to calculated. Each agent's local feature vectors are shared with their adjoining neighbors. We note it the weighted-edge between two agents $m$ and $n$. 
$\alpha _{m,n}$ gives the weight of the agent $m$ from $n$. The attention weight is measured by the following method.
A node $m$ summarizes the feature vector from its neighbor $N_m$, the summed vector $h_m$ will be.
For the sequential nodes in its neighbor $N_m$, this update of $h_m$ is unchanged. 
In the simulation, a fully connected network is adopted, and $N_m$ is the set of all other agents excluding $m$. 
The structure can, nevertheless, be integrated directly into any other adjacency relationship.

\begin{figure}[h]
    \centering
    \includegraphics[scale=0.45]{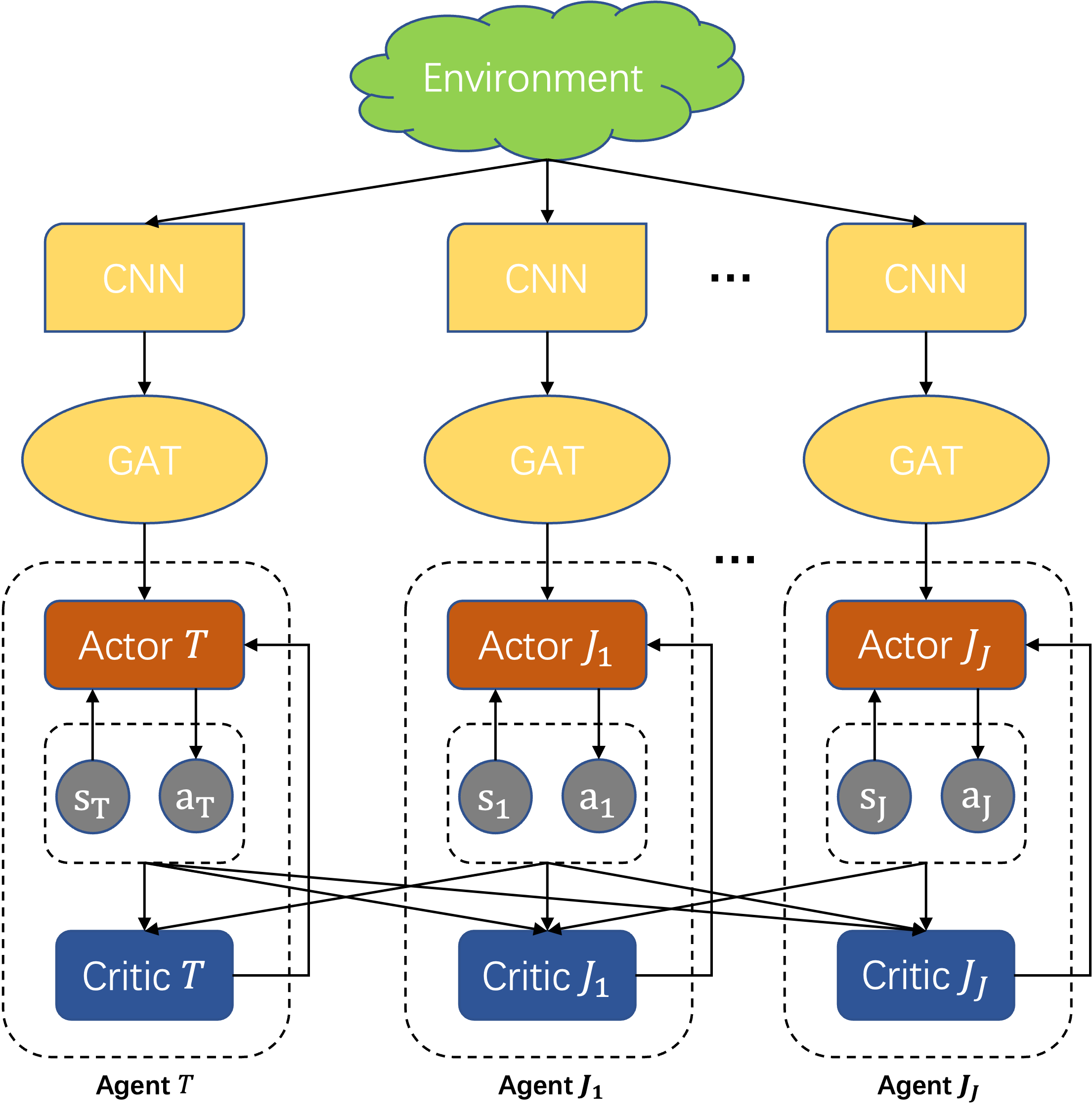}
    \caption{The architecture of the proposed algorithm.}
    \label{fig:my_label}
\end{figure}

\subsection{Multi-agent communication}
In a multi-agent setting, each agent's decision can be affected by each agent's joint actions in the network. We apply Markov games for modeling the multi-agent RL problem. It can be thought of as a multi-agent extension of MDPs.

The aggregate feature vector $h_m$ from GAT and the feature vector compose the input. We introduce the actor-critic algorithm to learn the best strategy. In path-planning tasks, the critical approach of policy gradient actors is usually better than DQN.

Each agent's network includes one actor and one critic. As the CM problem is a collaborative issue, the rewards are equivalent for each agent.
We note that the parameters of the actors as $t_1$, the parameters of the critics as $t_2$. ans the shared reward as $r(t, t_1)=\sum_{m=1}^k\pi (t_1)(t)$.

\subsection{Constant monitoring}
In this section, continuous monitoring issues are described. We assume the environment is a three-dimensional grid world $G$. Every unit in this environment will be monitored by agent. We associate the reward with each unit. Let $R_{k_t}$ be the relevant reward of a unit $u$ in $U$. 

We studied the learning effects when using grid maps, real world maps, and real world and grid maps. Assuming that $\beta$ is the negative decay in each agent. The reward is reset to $0$ or declines at a rate of $\beta$. If $m$ is monitored at the time of $t$, then the reward is $0$; otherwise, the reward is $\beta$.

Since $\beta >=0$, $R_{t_u}$ is a non-positive value. $R_{t_u}$ is a negative number of $R_{t_u}$, that is, a negative number related to the penalty of not viewing the unit $u$. The longer the interval of continuous observation, the greater the cumulative penalty. 
We set $R_max$ to the upper limit of the penalty, to prevent it from increasing indefinitely.

As Fig. 5 shows, consider that every agent obtains the same maximum coverage, a circle with a radius of $R$ centered on the agent. We consider two cases, one is that the agent can obtain global information in the whole environment, another is that the agent can only get local information.


Consistency monitoring aims to acquire the optimal policy $\pi$ which maximizes the sum reward for all agents over a finite length of time $T$.
This policy maps the state to each agent's actions. Each agent can choose one of its neighbors and stay in place.

As the reward is related to the agent’s historical location, the state must both include the agent’s location and the reward value of every unit in the present time slot. 
This guarantees Markov characteristics. 


However, as the state space is very large, we use RL to solve this problem.
By studying the effect of RL through the following methods, the diagram tends to be the low-resolution version of the learning environment. Besides, the grid map only consists of the location of the present one, the penalty value as well as the location of obstacles. As for the real world map, it is approximately the observations designated by those sensors decorated on the agents.

\begin{figure}[h]
    \centering
    \includegraphics[scale=0.6]{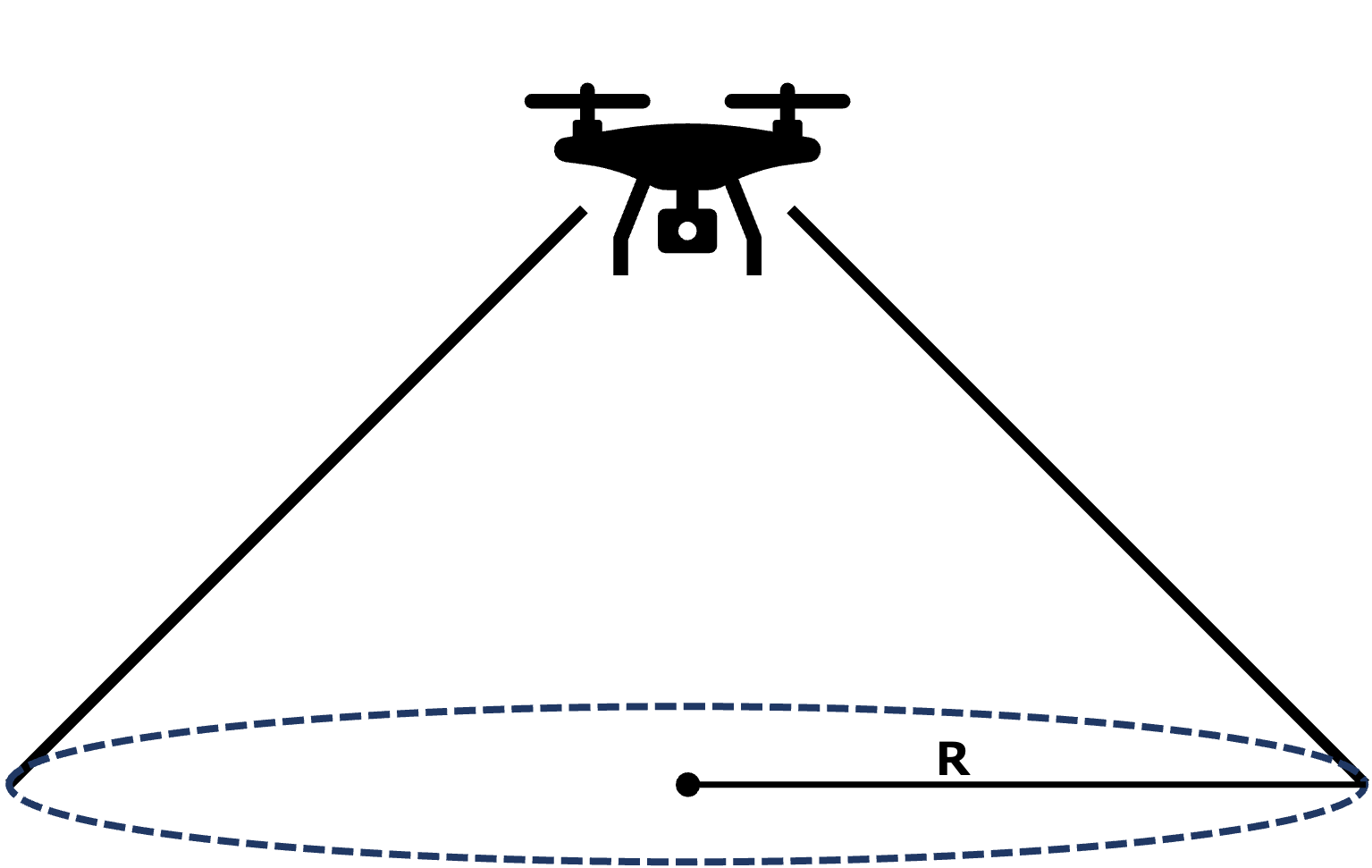}
    \caption{The maximum coverage of the UAV.}
    \label{fig:my_label}
\end{figure}

\section{Simulation Results}
To quantify our proposed algorithm, we consider two performance metrics: the MOS and the offloading rate, we perform simulation experiments and evaluate the performance of the M-AGCDRL algorithm. 
First, we present the simulation scenario. Then, the effects of various indicators of the simulation are analyzed. 

\subsection{simulation scene}
As it is necessary to obtain all the vehicles' position at the beginning of each cycle, the simulation tool SUMO \cite{dlr124092} is applied to generate vehicle trace. Specially, we implement two scenarios: a $10X10$ synthetic traffic grid and a real-world traffic network from Sapporo city. Concrete IoV trace data is generated by the help of the simulation tool SUMO. Specifically, we intercepted a map of Sapporo City where lanes are given a speed limit of $60km/h$.

With the generated vehicle trace, we implement our algorithms on TensorFlow using Python 3. Table II shows the parameters of the environment and base stations. Denote a UAV-assisted wireless network is made up of $M$ vehicles and $N$ UAVs. The bandwidth of the BS is $B=10MHz$. 
We consider there are at most $100$ vehicles moving on the lanes.

Meanwhile, we set the parameter of Q-learning as Table III. The learning rate is $lr$ = 0.01. The reward decay is $\gamma  = 0.9$. The threshold $t = 0.1$. Then set $ep=5000$, which means we simulate 5000 episodes.

\begin{table}[h]
\centering
\caption{Parameters of Simulation}
\begin{tabular}{|l|l|}
\hline
Parameter & Value \\ \hline
B         & 10MHz \\ \hline
$p_{t_i}$ & 4kW   \\ \hline
$p_{c_i}$ & 40W   \\ \hline
N         & 5     \\ \hline
M         & 100   \\ \hline
lr        & 0.01  \\ \hline
$\gamma $ & 0.9   \\ \hline
t         & 0.1   \\ \hline
ep        & 5000  \\ \hline
\end{tabular}
\end{table}

\subsection{Numerical Results}

\begin{figure*}[h]
    \centering
    \includegraphics[scale=0.123]{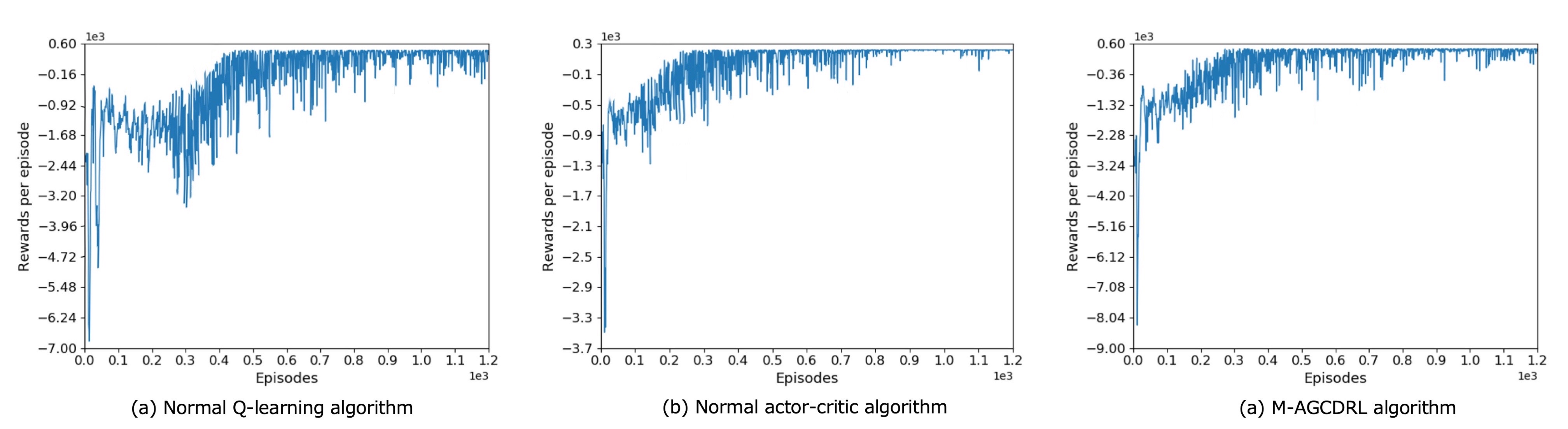}
    \caption{The total rewards with 10 MEC server of each episode versus algorithms under grid world map.}
    \label{Fig. 1: UAV-enabled wireless communications architecture}
\end{figure*}

\begin{figure*}[h]
    \centering
    \includegraphics[scale=0.123]{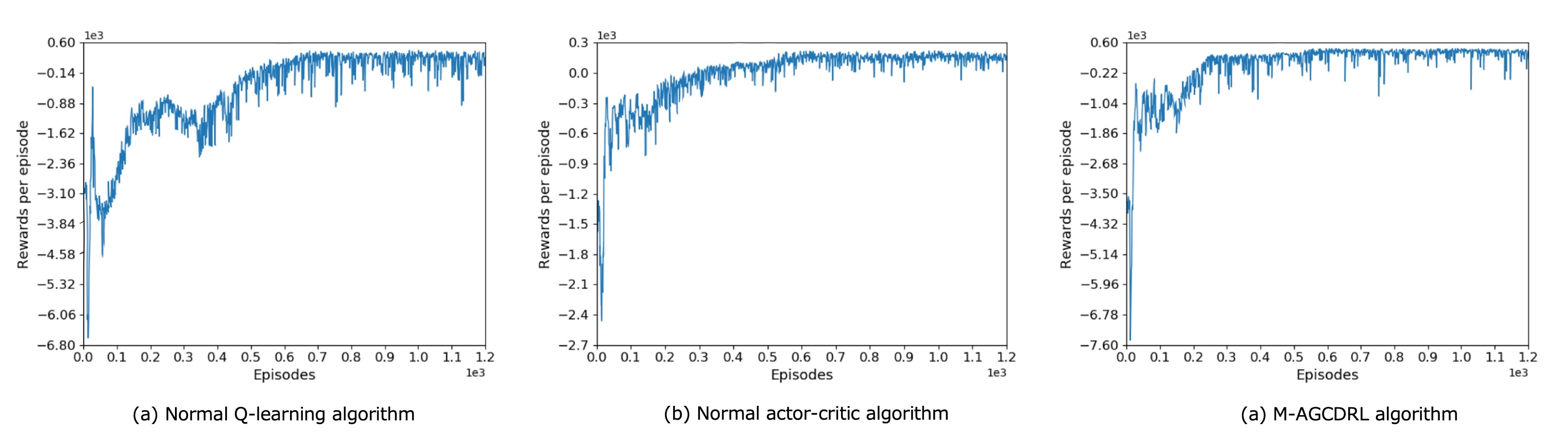}
    \caption{The total rewards with 10 MEC server of each episode versus algorithms under real world map.}
    \label{Fig. 1: UAV-enabled wireless communications architecture}
\end{figure*}


Fig. 6 and Fig. 7 show the convergence speed of the proposed algorithm, the ordinary actor-critic algorithm, as well as the ordinary Q-learning algorithm under grid world map and real world map, respectively. As we can see from the figures, the total reward for each of the 1,200 episodes in the learning phase fluctuates sharply and relatively little during the first few hundred episodes, then stabilize at a comparatively high level. 
Once 10,000 experiences are saved, the learning phase of the proposed algorithm starts updating the parameters. As a result, the total reward per episode fluctuates dramatically at the beginning then gradually optimized. 


Fig. 8 illustrates the impact of the number of MEC devices performing computational offloading. With the increase in the number of UAVs, the number of UAVs performing computational offloading also increases, and all of them tend to be stable at the last. 
Also, we can find that the proposed method always performs better.

\begin{figure}[H]
    \centering
    \includegraphics[scale=0.35]{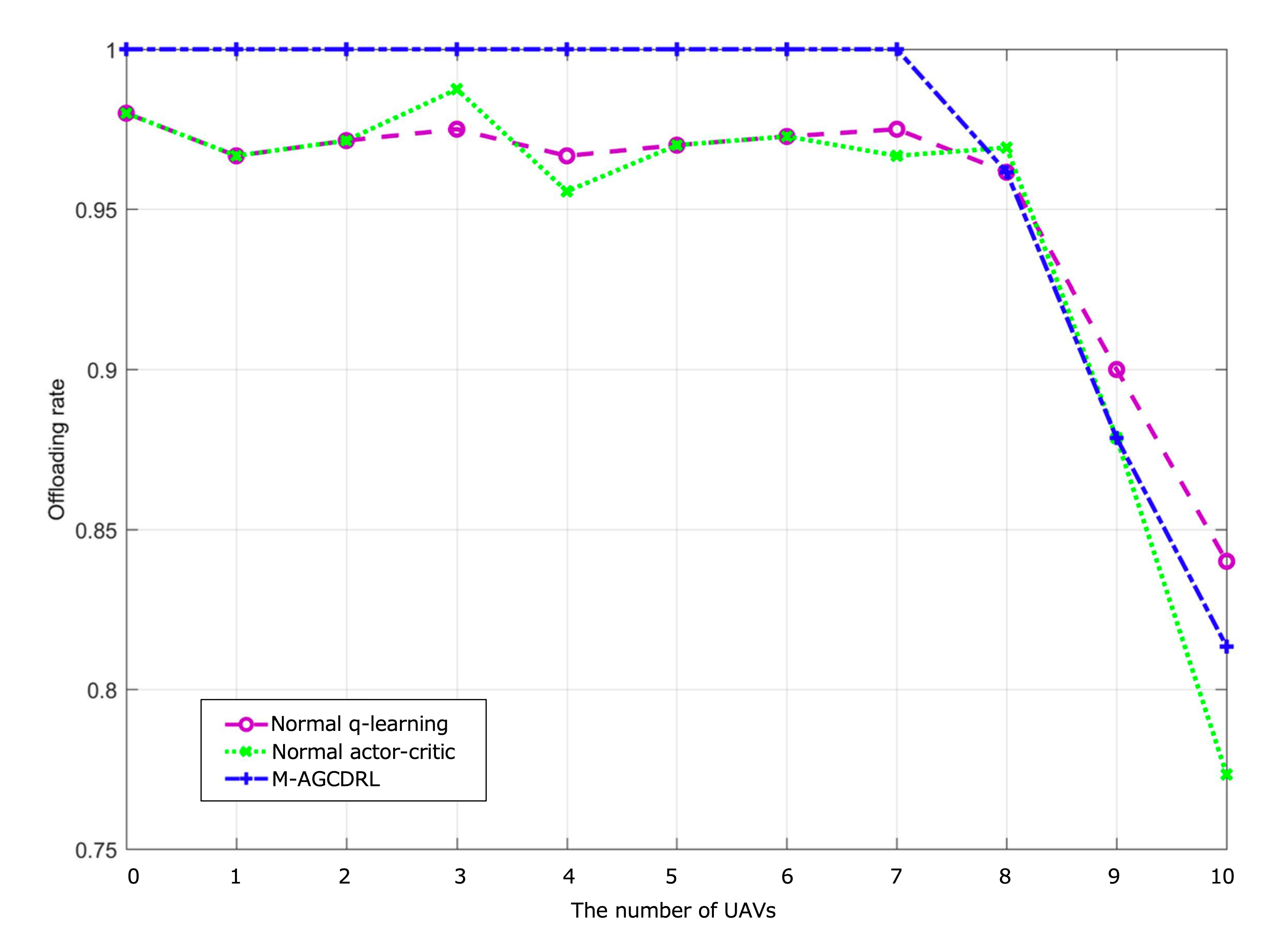}
    \caption{The number of UAVs performing computation offloading versus different method.}
    \label{fig:my_label}
\end{figure}

Fig. 9 shows offloading performance as the number of UAVs gets greater. 
From the figure, we can see that at the very beginning when the number of UAVs is not very large, the the offloading tends to perform better because the number of channels to the UAVs also increase as the number UAVs increases. 
As the limited computing resources, if extravagant UAVs are accessing, the delay will also rise. In this way, the system finally get to stable.

\begin{figure}[H]
    \centering
    \includegraphics[scale=0.62]{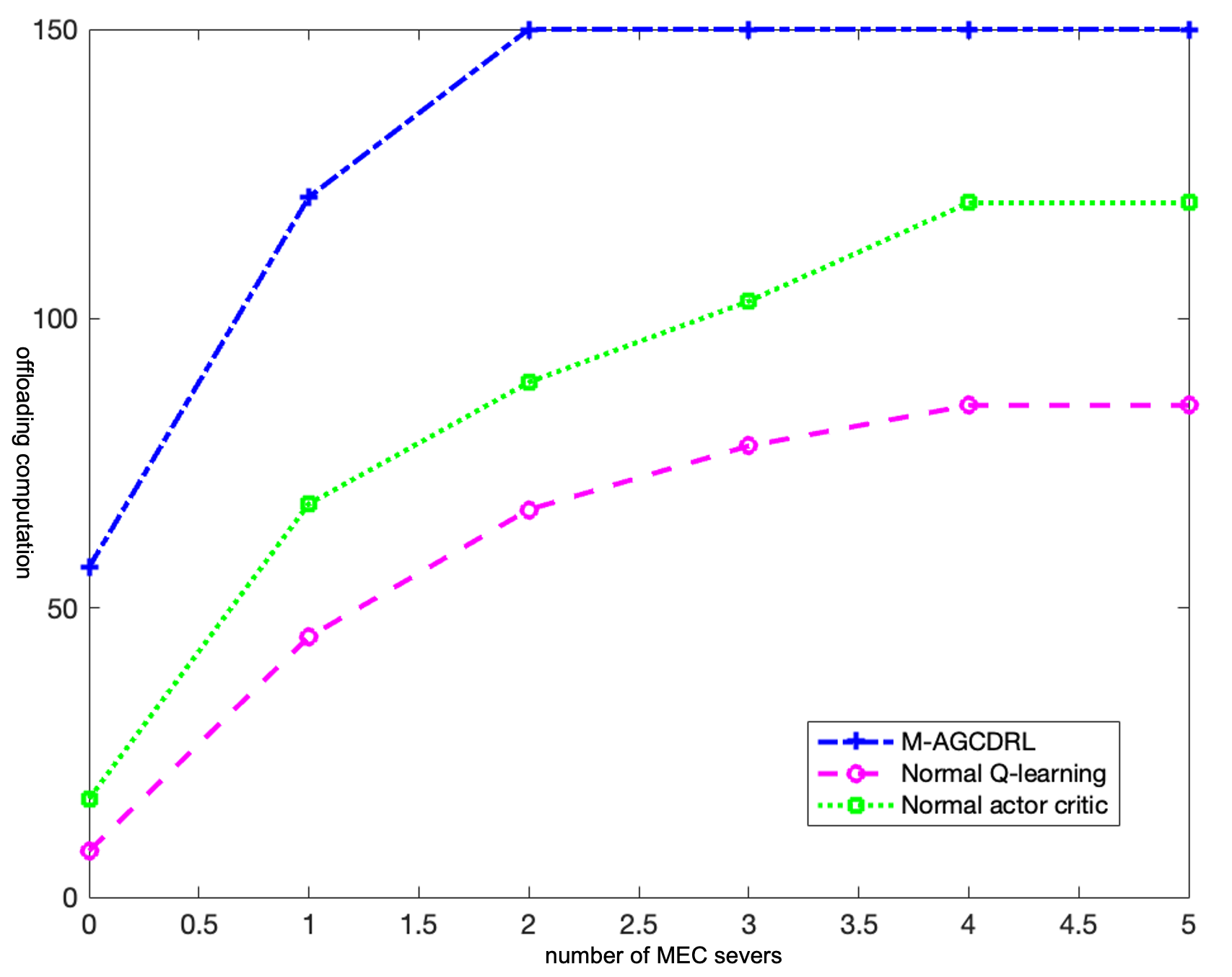}
    \caption{The number of UAVs performing computation offloading as MEC servers increase.}
    \label{fig:my_label}
\end{figure}

Fig. 10 shows the average QoS, in other words, the MOS under the proposed algorithm, the common actor-critic algorithm and the common Q-learning algorithm. As available computing resources volume increases, the MOS by our proposed algorithm and the comparison algorithms increases because of the increase in the volume of resources assigned to the drivers on the lane.

\begin{figure}[H]
    \centering
    \includegraphics[scale=0.22]{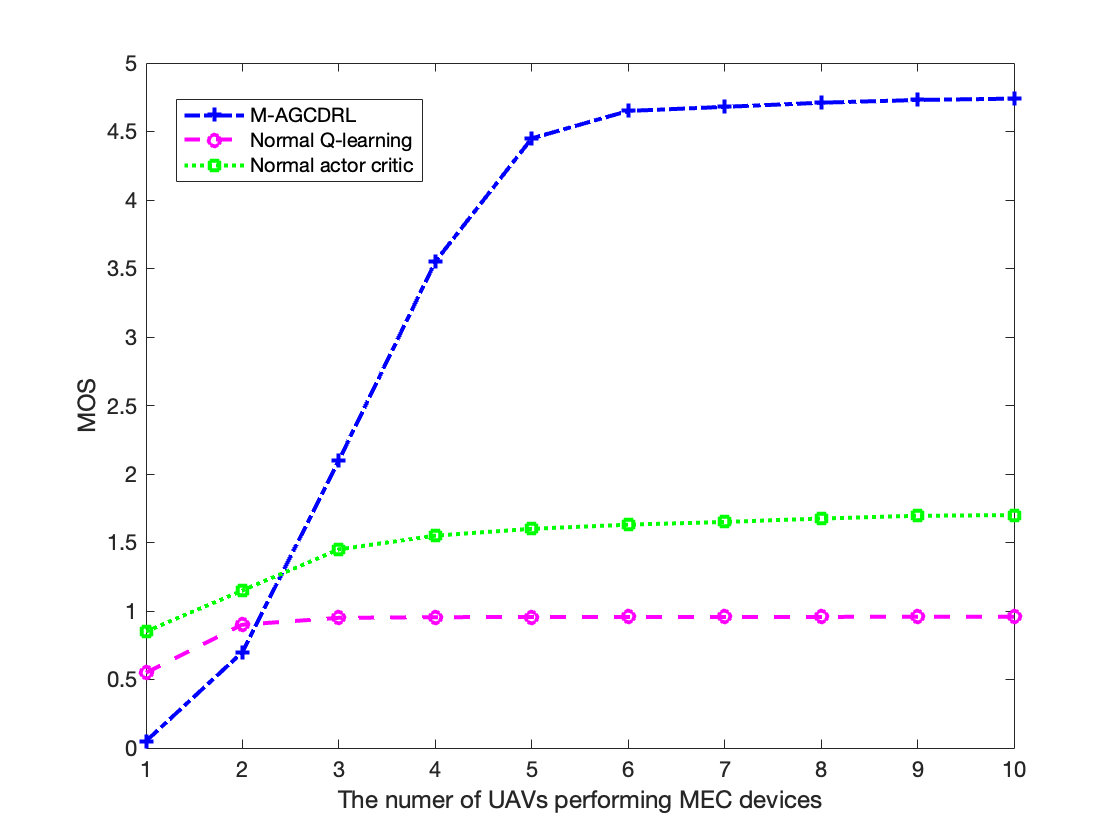}
    \caption{MOS with UAVs increasing versus algorithm.}
    \label{fig:my_label}
\end{figure}

\section{Conclusion}
Given the tremendous promise of 5G, it is clear that facilitating circular communication in 5G cellular networks based on MEC offers significant advantages in terms of scalability, reliability, mobility support and packet latency.

In this paper, we have we present the M-AGCDRL algorithm to QoE-driven MEC schedule problem. 
We combine the graph convolutional networks and actor-critic architecture to deal with the modeled MDPs. 
Through the simulation, we can see that our proposed algorithm could offloading rate compared with the normal q-learning algorithm and actor critic algorithm, which could lead to a better QoE for drivers on the lane. 
In addition, our algorithm is computationally efficient and can converge more quickly than the common algorithms.
Through the study, the MEC-assisted wireless network from fundamental theory to practical implementation can be investigated in a more comprehensive way.

\section*{Acknowledgments}
This work is partially supported by JSPS KAKENHI Grant Numbers JP19K20250, JP20F20080, and JP20H04174, Leading Initiative for Excellent Young Researchers (LEADER), MEXT, Japan, and JST, PRESTO Grant Number JPMJPR21P3, Japan. Mianxiong Dong is the corresponding author.


%


\bibliographystyle{IEEEtran}
\bibliography{IEEEabrv, UAV}

\newpage

\section{Biography Section}
\begin{IEEEbiography}[{\includegraphics[width=1in,clip,keepaspectratio]{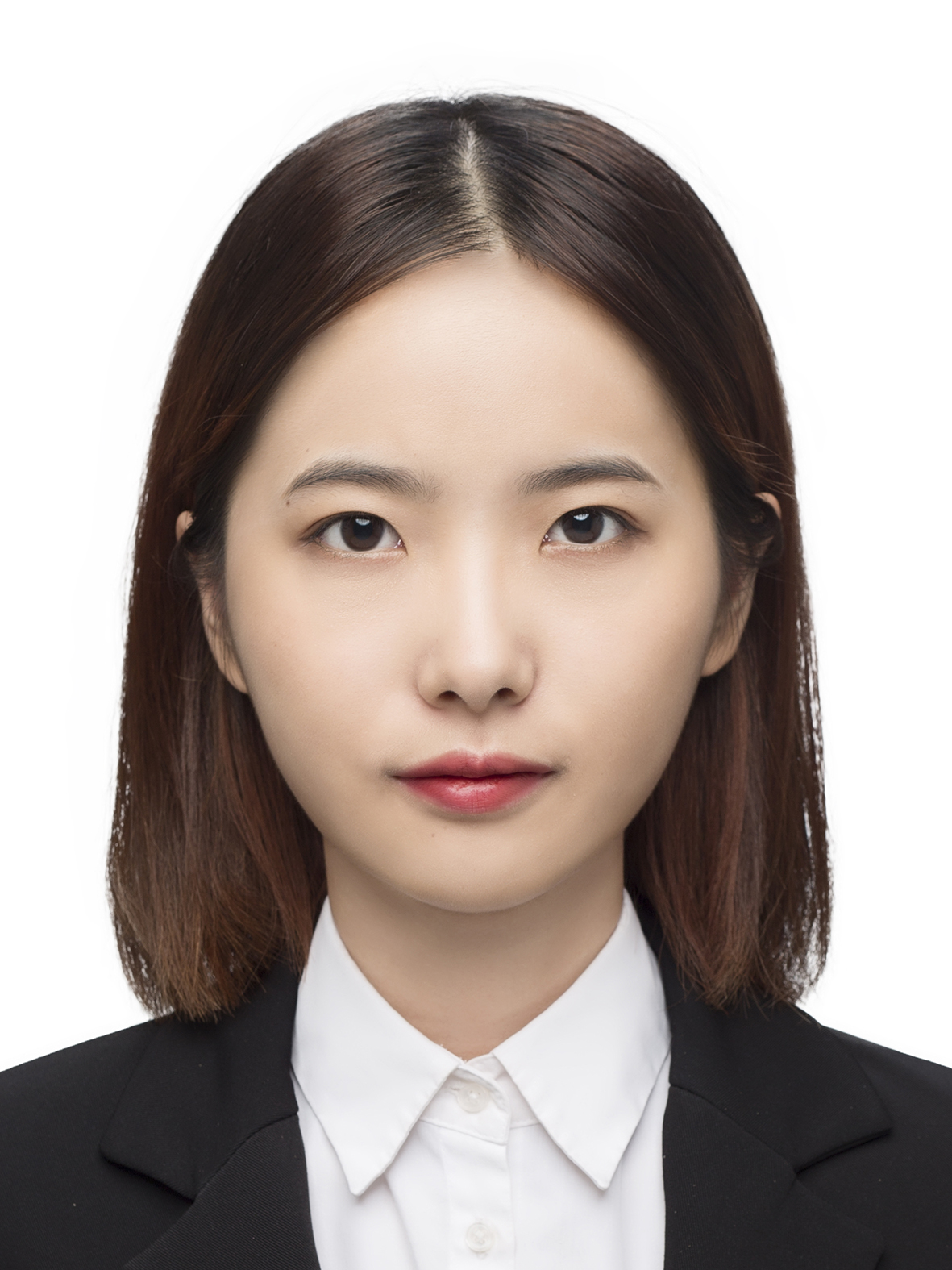}}]{Xiaoyu Dai}
received the B.E. degree in School of Cyber Science and Engineering from Wuhan University, China and was a research student with ENeS lab at Muroran Institute of Technology, Japan, the same year. She is currently pursuing the M.S. degree in Muroran Institute of Technology, Japan.
\end{IEEEbiography}
 
\begin{IEEEbiography}[{\includegraphics[width=1in,clip,keepaspectratio]{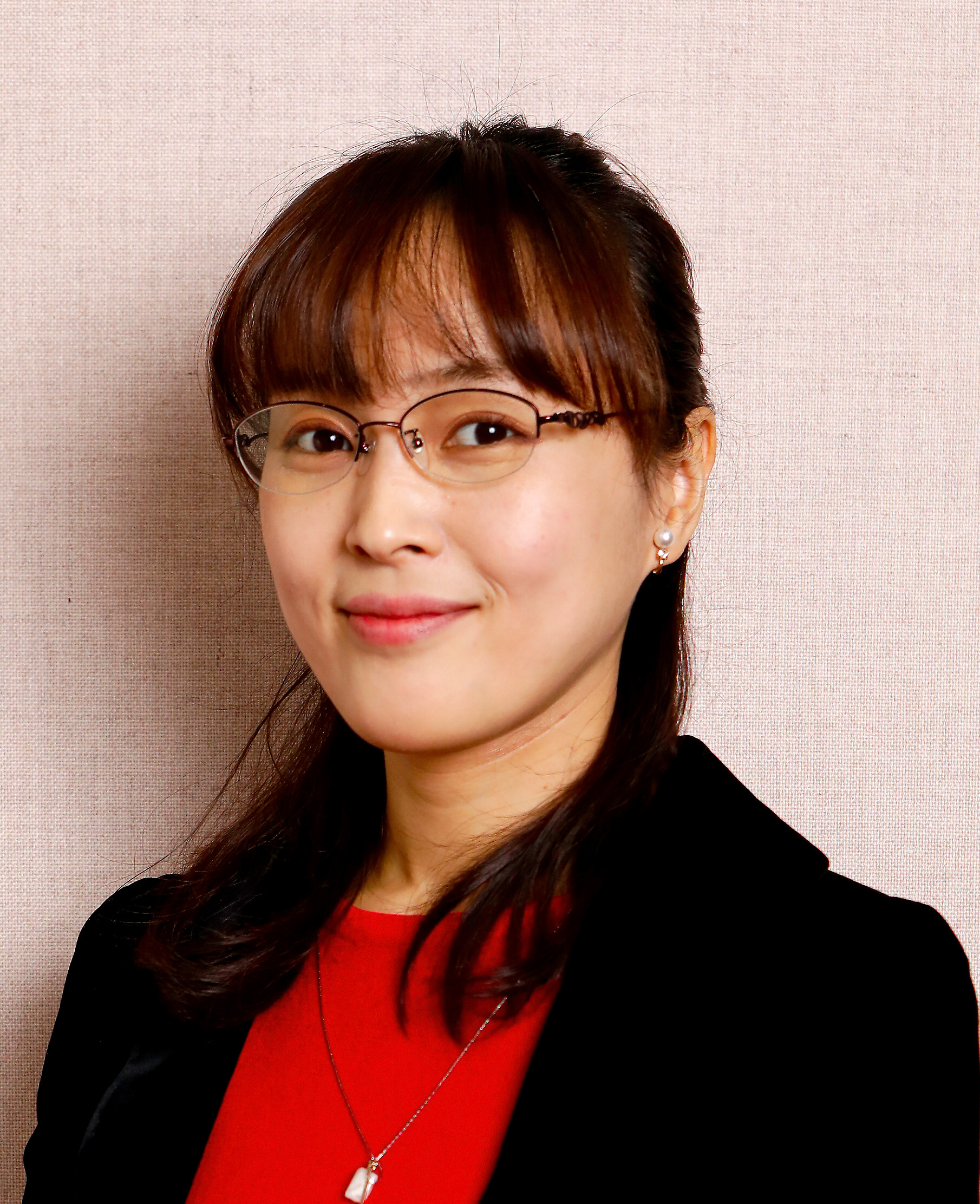}}]{Kaoru Ota}    
  was born in Aizu-Wakamatsu, Japan. She received M.S. degree in Computer Science from Oklahoma State University, the USA in 2008, B.S. and Ph.D. degrees in Computer Science and Engineering from The University of Aizu, Japan in 2006, 2012, respectively. Kaoru is currently an Associate Professor and Ministry of Education, Culture, Sports, Science and Technology (MEXT) Excellent Young Researcher with the Department of Sciences and Informatics, Muroran Institute of Technology, Japan. From March 2010 to March 2011, she was a visiting scholar at the University of Waterloo, Canada. Also, she was a Japan Society of the Promotion of Science (JSPS) research fellow at Tohoku University, Japan from April 2012 to April 2013. Kaoru is the recipient of IEEE TCSC Early Career Award 2017, The 13th IEEE ComSoc Asia-Pacific Young Researcher Award 2018, 2020 N2Women: Rising Stars in Computer Networking and Communications, 2020 KDDI Foundation Encouragement Award, and 2021 IEEE Sapporo Young Professionals Best Researcher Award. She is Clarivate Analytics 2019, 2021 Highly Cited Researcher (Web of Science) and is selected as JST-PRESTO researcher in 2021, Fellow of EAJ in 2022.
 \end{IEEEbiography}
 
 \begin{IEEEbiography}[{\includegraphics[width=1in,clip,keepaspectratio]{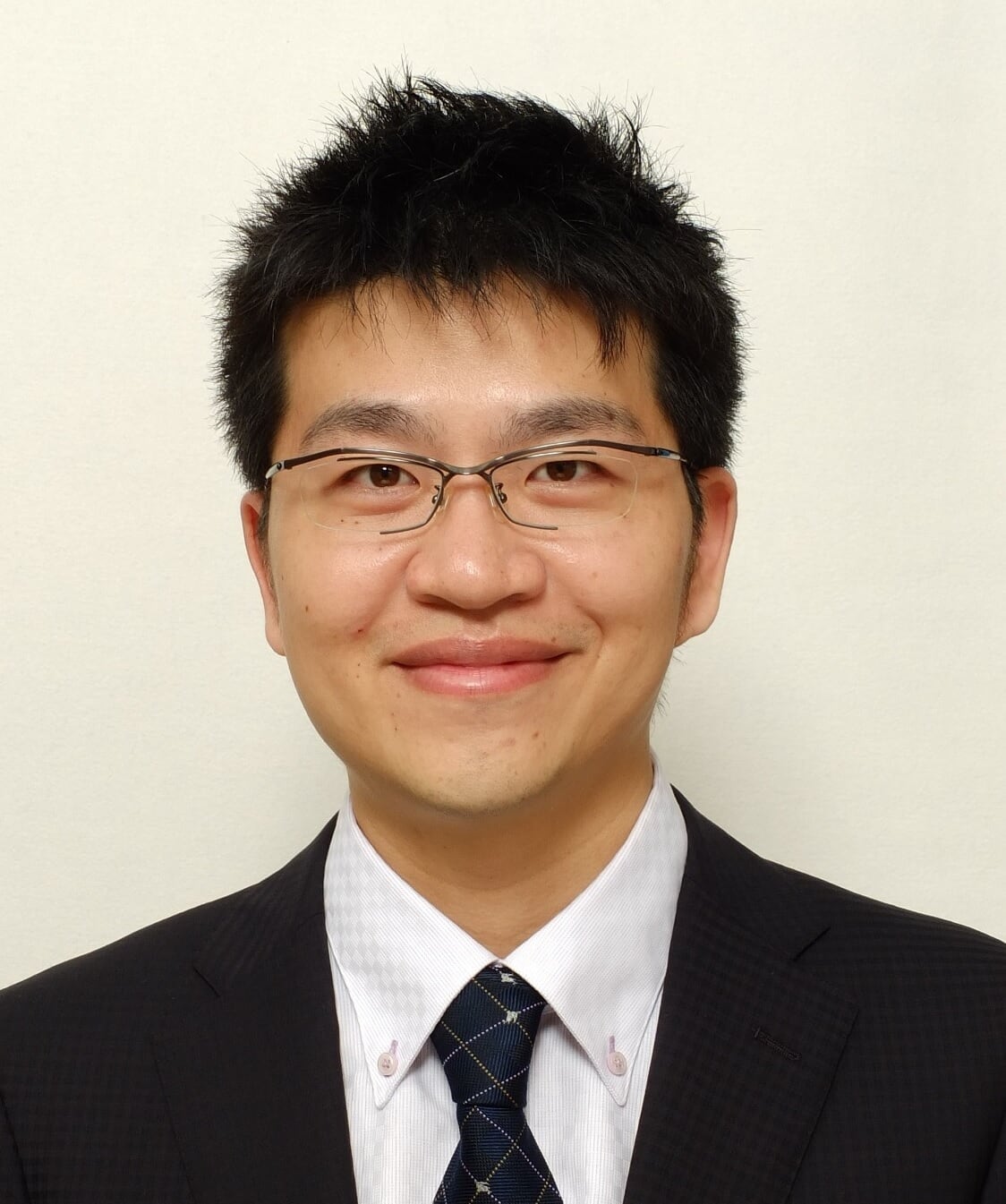}}]{Mianxiong Dong} received B.S., M.S. and Ph.D. in Computer Science and Engineering from The University of Aizu, Japan. He is the Vice President and Professor of Muroran Institute of Technology, Japan. He was a JSPS Research Fellow with School of Computer Science and Engineering, The University of Aizu, Japan and was a visiting scholar with BBCR group at the University of Waterloo, Canada supported by JSPS Excellent Young Researcher Overseas Visit Program from April 2010 to August 2011. Dr. Dong was selected as a Foreigner Research Fellow (a total of 3 recipients all over Japan) by NEC C\&C Foundation in 2011. He is the recipient of The 12th IEEE ComSoc Asia-Pacific Young Researcher Award 2017, Funai Research Award 2018, NISTEP Researcher 2018 (one of only 11 people in Japan) in recognition of significant contributions in science and technology, The Young Scientists’ Award from MEXT in 2021, SUEMATSU-Yasuharu Award from IEIEC in 2021, IEEE TCSC Middle Career Award in 2021. He is Clarivate Analytics 2019, 2021 Highly Cited Researcher (Web of Science) and Foreign Fellow of EAJ.
 \end{IEEEbiography}

\vfill

\end{document}